\documentclass[
]{ceurart}

\usepackage{listings}
\begin{document}

\copyrightyear{2025}
\copyrightclause{Copyright for this paper by its authors.
  Use permitted under Creative Commons License Attribution 4.0
  International (CC BY 4.0).}

\conference{CLEF 2025 Working Notes, 9 -- 12 September 2025, Madrid, Spain}

\title{Are Smaller Open-Weight LLMs Closing the Gap to Proprietary Models for Biomedical Question Answering?}

\title[mode=sub]{Notebook for the BioASQ Lab at CLEF 2025}





\author[1]{Damian Stachura}[%
email=damian.stachura@evidenceprime.com
]
\address[1]{Evidence Prime, Krakow, Poland}

\author[1]{Joanna Konieczna}[
email=joanna.konieczna@evidenceprime.com
]

\author[1]{Artur Nowak}[%
email=artur.nowak@evidenceprime.com
]


\begin{abstract}
  Open-weight versions of large language models (LLMs) are rapidly advancing, with state-of-the-art models like DeepSeek-V3 now performing comparably to proprietary LLMs. This progression raises the question of whether small open-weight LLMs are capable of effectively replacing larger closed-source models. We are particularly interested in the context of biomedical question-answering, a domain we explored by participating in Task 13B Phase B of the BioASQ challenge. In this work, we compare several open-weight models against top-performing systems such as GPT-4o, GPT-4.1, Claude 3.5 Sonnet, and Claude 3.7 Sonnet. To enhance question answering capabilities, we use various techniques including retrieving the most relevant snippets based on embedding distance, in-context learning, and structured outputs. For certain submissions, we utilize ensemble approaches to leverage the diverse outputs generated by different models for exact-answer questions. Our results demonstrate that open-weight LLMs are comparable to proprietary ones. In some instances, open-weight LLMs even surpassed their closed counterparts, particularly when ensembling strategies were applied. All code is publicly available at \href{https://github.com/evidenceprime/BioASQ-13b}{https://github.com/evidenceprime/BioASQ-13b}.
\end{abstract}

\begin{keywords}
  Biomedical Question Answering \sep
  Large Language Models \sep
  Zero-Shot Prompting \sep
  Few-Shot Prompting \sep
  In-Context Learning \sep
  GPT-4 \sep
  Claude \sep
  Open-Weight LLM \sep
  Ensembling
\end{keywords}

\maketitle

\section{Introduction}
In question answering tasks, access to domain-specific knowledge can significantly enhance response quality, particularly when answers need to be grounded in provided supplementary materials.

The BioASQ Challenge \cite{BioASQ2025overview} exemplifies such a task in biomedical domain. In this challenge, participating systems are provided with relevant biomedical papers from the PubMed database. These materials can then be leveraged to generate high-quality responses to the posed questions. The questions themselves span four distinct types: yes/no, factoid, list, and summary questions.

Over the years, various approaches have been employed for question answering tasks in the BioASQ Challenge. In its earliest editions, classic methods were applied, such as BM25, which ranked retrieved documents based on their relevance to a question. Additionally, researchers applied similarity algorithms using vector embeddings, algorithms based on linguistic annotations of texts, and early deep learning models. For an extended period, BERT-based solutions, notably BioBERT \cite{biobert} and PubMedBERT \cite{pubmedbert}, dominated these question answering tasks. Sequence-to-sequence models like T5 \cite{t5} also proved to be useful.

However, since the global emergence of ChatGPT \cite{brown2020languagemodelsfewshotlearners, ChatGPT} in 2022, large language models have significantly reshaped the competitive landscape of question answering. Proprietary models, including OpenAI's offerings, Gemini \cite{geminiteam2025geminifamilyhighlycapable}, and Claude \cite{anthropic2023claude}, initially dominated this field, fostering the belief that creating robust LLMs requires a massive financial investment.

A significant shift began in 2024 and 2025, with a growing number of organizations publicly releasing models featuring open weights and permissive licenses. Today, large open-weight models like Llama 3-405B \cite{grattafiori2024llama3herdmodels}, DeepSeek-V3 \cite{deepseekai2025deepseekv3technicalreport}, and Qwen3-235B-A22B \cite{qwen3} are proving capable of challenging even the best proprietary models, as evidenced by platforms like the LM Arena \cite{chiang2024chatbotarenaopenplatform}. Perhaps even more impactful are the smaller open-weight models, which can run on consumer-grade machines and demonstrate impressive competitiveness in tasks requiring access to domain-specific knowledge, especially within a retrieval-augmented generation (RAG) setup. 

This year marks the 13th edition of the BioASQ Challenge, and we participated in Task 13B, Phase B \cite{BioASQ2025task13bSynergy}. Our primary objective was to investigate whether relatively small LLMs, primarily those up to 14 billion parameters, could effectively compete with more powerful proprietary models in this biomedical question-answering context. To achieve this, we explored multiple strategies for enhancing our results. We utilized in-context learning by leveraging the provided database of questions from previous BioASQ challenge editions. Additionally, we used similarity algorithms using vector embeddings to select a pertinent subset of snippets from the provided PubMed articles for each question. This paper details the results from four question batches of Task 13B Phase B and discusses our conclusions.

\section{Methodology\label{methods}}
We experimented with numerous techniques to optimize performance in biomedical question answering. The successful approaches implemented in our solutions are outlined below.

\subsection{Best Snippets Selection}
For our submissions, we selected the 10 best-matching snippets from the provided PubMedQA articles. Our team experimented with varying snippet counts, validating them against datasets from previous BioASQ challenge versions. This process led us to select 10 snippets as the optimal number, as it consistently produced the most robust results. We utilized the sentence-transformers library \cite{reimers-2019-sentence-bert} with the \texttt{nomic-embed-text-v1} \cite{nussbaum2024nomic} model. Our approach involved computing embeddings for all snippets and the question. Subsequently, we calculated the cosine similarity between each (snippet, question) pair to identify options with the highest similarity. Finally, these selected snippets were provided to the model, ordered from most to least similar.

\subsection{In-Context Learning}
Research has demonstrated that in-context learning enhances the performance of language models in diverse applications \cite{brown2020languagemodelsfewshotlearners, min-etal-2022-rethinking}. We investigated how different models performed with in-context learning, sourcing examples from previous BioASQ challenge editions. We used Qdrant \cite{qdrantdb}, a vector database, into which we inserted computed embeddings for all previous questions combined with their 10 best snippets. Subsequently, for each new question, we queried the database for the most similar elements, following the approach presented in \cite{Rubin2021LearningTR}. Experimentally, we determined that 3 examples were optimal for factoid and list questions, while a zero-shot approach was used for yes/no and summary type questions.

\subsection{Prompts}
For all question types, we utilized hand-crafted prompts. As noted previously, a zero-shot prompting approach was employed for yes/no and summary questions, given empirical observations that few-shot prompting detrimentally affected performance for these specific categories. Conversely, for factoid and list questions, few-shot prompting demonstrated clear benefits. Accordingly, we implemented a 3-shot prompting strategy for these questions, based on insights gained through comprehensive experimentation. The system prompts are detailed in Table~\ref{tab:sysprompts}, and the actual prompts guiding the models to generate answers for all question types are presented in Table~\ref{tab:prompts}. We also briefly experimented with DSPy \cite{khattab2024dspy} for the automated generation of prompts based on predefined input and output schemas in batch 2. However, responses achieved using DSPy were slightly worse than those achieved with our hand-crafted prompts. In the future, we plan to investigate whether automatic prompt optimization can help by creating model-specific prompts.

\begin{table*}
  \caption{System prompts for exact answers and ideal answers questions}
  \label{tab:sysprompts}
  \begin{tabular}{l|p{3in}}
    \toprule
    Question type & System prompt \\
    \midrule
    Exact answer & You are a biomedical AI expert specializing in question answering, research, and entity extraction. \\
    \midrule
    Ideal answer & You are an expert in the medical texts summarization. Answer the given question with a single paragraph text and your answer should be based on the provided context snippets. You should generate your response in at most 2-3 sentences (30-50 words). \\
    \bottomrule
  \end{tabular}
\end{table*}

\begin{table*}
  \caption{Prompts for all types of questions in the Task 13B Phase B}
  \label{tab:prompts}
  \begin{tabular}{l|p{3.5in}}
    \toprule
    Question type & Prompt \\
    \midrule
    Yes/No & Given only the following SNIPPETS and QUESTION, answer the QUESTION only with ’Yes’ or ’No’. \\
    \midrule
    Factoid & Extract key biomedical entities **strictly using the provided SNIPPETS** to answer the QUESTION. List **1 to 5** of the most relevant entities, ranked by confidence. **Never exceed 5 entities.** If more exist, return only the top 5. Prefer concise entities and **remove redundant or longer variants** of the same term. If no relevant entities exist, return `None.`. \\
    \midrule
    List & Extract key biomedical entities **strictly using the provided SNIPPETS** to answer the QUESTION. List **1 to 5** of the most relevant entities. Prefer concise entities and **remove redundant or longer variants** of the same term. If no relevant entities exist, return `None`. \\
    \midrule
    Summary & Answer the QUESTION by returning a single paragraph sized text (use max 50 words) ideally summarizing only the most relevant information in the SNIPPETS. \\
  \end{tabular}
\end{table*}

\subsection{Structured Outputs}
We opted to use structured outputs to facilitate the extraction of LLM results in a predefined format. We defined a JSON schema for response formatting, and subsequently followed a context-free grammar (CFG) approach for it. We used CFG implementations introduced by model providers or accessed via external libraries like Outlines \cite{outlines} to guide the token sampling process. This methodology ensures that the generated tokens adhere strictly to the schema, eliminating the need for complex regex-based extraction from the model response.

\subsection{Models}
In our study, we utilized various LLMs, drawing from both open-weight and closed options. Our primary focus was on relatively smaller open-weight models, specifically those with up to 14 billion parameters, such as Phi-4 \cite{phi4techreport}, Gemma-3-12B \cite{gemma3}, Qwen2.5 14B \cite{qwen2025qwen25technicalreport}, and Meditron Phi-4 14B \cite{meditron}. For a third batch of experiments, we expanded our testing to include quantized versions of Gemma3-27B \cite{gemma3} and Mistral3-24B \cite{mistral3_1}. Although we briefly attempted to use HuatuoGPT-o1 \cite{huatuogpto1}, our limited exploration of reasoning models meant we did not achieve strong results with it. We also incorporated several of the newest closed-source models, including recent generations of GPT (GPT-4o \cite{openai2024gpt4ocard}, GPT-4.1 \cite{gpt4_1}) and Claude (Claude Sonnet 3.5 \cite{claude3_5}, Claude Sonnet 3.7 \cite{claude3_7}).

\subsubsection{Quantized Models}
For batches 1-3, our experiments used open-weight models quantized to 4-bit. However, in the final batch, we proposed solutions based on the full, unquantized versions of these models. Interestingly, in both setups, the open-weight models proved to be competitive with the closed alternatives.

\subsection{Ensembling Method}
Ensembling methods are widely recognized as beneficial when combining responses from multiple weaker models to achieve a single, stronger result. Various techniques exist for this purpose, including majority voting, confidence scoring, and aggregation.

\begin{figure}
  \centering
  \includegraphics[width=\linewidth]{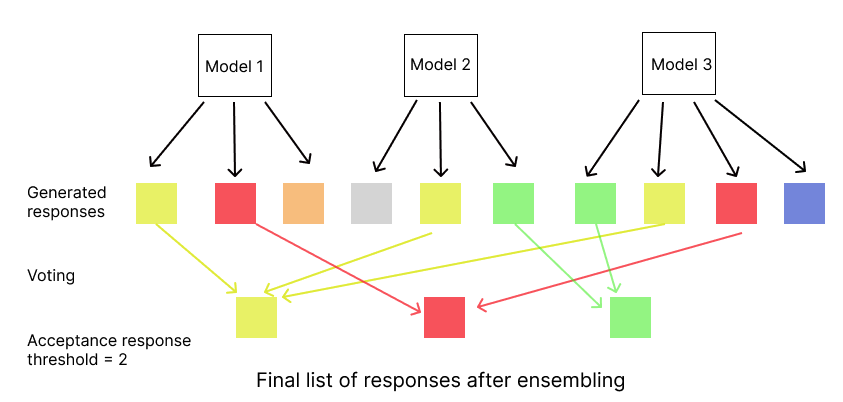}
  \caption{Ensembling strategy for factoid and list type of questions}
  \label{fig:ensamble}
\end{figure}

For yes/no questions, we applied a straightforward majority voting approach, where the final answer was determined by the option chosen by the most models. For factoid and list questions, we developed a more sophisticated aggregation method. This involved collecting responses from all models and calculating the frequency of each distinct response. The most frequent response was then selected, provided its number of appearances exceeded a predefined threshold. The process is visualized on Figure~\ref{fig:ensamble}. For factoid questions, we limited the output list to a maximum of five best responses, as specified in the rules for this question type.

For later batches, we incorporated different classes of LLMs into our ensembling strategy. This approach leveraged the observed benefits of integrating the diverse characteristics of various LLM families such as Phi, Qwen, Mistral, and Gemma. As emphasized by Jiang et al. \cite{jiang2023llmblenderensemblinglargelanguage}, different LLMs, trained on varied data and architectures, inherently exhibit unique strengths that can be synergistic in an ensemble.

\section{Results}
We participated solely in Task 13B, Phase B, of the BioASQ challenge \cite{BioASQ2025overview}\cite{BioASQ2025task13bSynergy}. This task encompasses four types of questions: yes/no, factoid, and list questions, which are evaluated by matching exact answers provided by the challenge organizers. In addition, there are summary questions that require an ideal and free-form summary as a response, scored through automatic metrics and manual reviews. Our approach involved testing multiple techniques and models across published question batches, leading to distinct strategies for each.

\subsection{System Definitions\label{systemdef}}
For each submission, the rules detailed in \ref{methods} were applied. The models used to generate responses in each submission, grouped by system name from 1 to 5, are presented below. The results for these systems in all batches are summarized in Table~\ref{tab:yesno}, Table~\ref{tab:factoid}, Table~\ref{tab:list}, and Table~\ref{tab:summary}. A detailed specification of each submitted system follows:
\newline

\textbf{Batch 1:}
\begin{itemize}
    \item EP-1: Phi-4
    \item EP-2: HuatuoGPT-o1 8B
    \item EP-3: Qwen2.5-14B
    \item EP-4: GPT-4o
    \item EP-5: Claude 3.5 Sonnet
\end{itemize}

\textbf{Batch 2:}
\begin{itemize}
    \item EP-1: Ensemble - Gemma-3-12B, Qwen2.5-14B, Phi-4, GPT-4o, Claude 3.5 Sonnet
    \item EP-2: Claude 3.5 Sonnet
    \item EP-3: Phi-4
    \item EP-4: Phi-4 + DSPy prompt (only for factoid questions, without prompt optimization)
    \item EP-5: Qwen2.5-14B + DSPy prompt (only for factoid questions, without prompt optimization)
\end{itemize}

\textbf{Batch 3:}
\begin{itemize}
    \item EP-1: Ensemble - Mistral-Small-3.1-24B, Gemma-3-12b, Gemma-3-27b, Qwen2.5-14B, Phi-4
    \item EP-2: Ensemble - GPT-4o, GPT-4.1, Claude 3.5 Sonnet
    \item EP-3: GPT-4.1
    \item EP-4: Phi-4
\end{itemize}


\textbf{Batch 4:}
\begin{itemize}
    \item EP-1: Ensemble - GPT-4.1, GPT-4o, Claude 3.5 Sonnet, Claude 3.7 Sonnet
    \item EP-2: Ensemble - Gemma-3-12B, Qwen2.5-14B, Meditron3-Phi4-14B, Phi-4
    \item EP-3: Ensemble - Qwen2.5-14B, Meditron3-Phi4-14B, Phi-4, GPT-4.1, GPT-4o, Claude 3.5 Sonnet, Claude 3.7 Sonnet
    \item EP-4: Ensemble - Qwen2.5-14B, Phi-4, GPT-4.1, GPT-4o, Claude 3.7 Sonnet
    \item EP-5: GPT-4.1
\end{itemize}

\subsection{Exact Answers}
We used distinct answering strategies for each batch, as detailed in \ref{systemdef}. For batch 1, our primary focus was on assessing the performance of individual models within each system. In batches 2 and 3, we also incorporated ensembling techniques, specifically involving combinations of open-weight and selected closed models. It is important to note that for these batches, we only used quantized versions of open models. Finally, for the last batch, we conducted a comparative analysis between full open-weight models and proprietary models.

The most advantageous approach for yes/no questions was difficult to determine. Proprietary models achieved the best results in batches 1 and 4, while open-weight models dominated in the remaining. It is shown in Table~\ref{tab:yesno}. The quality of the provided context appears to be a critical factor, with both model types demonstrating sufficient capability to extract key information pertinent to the question.

List-based questions demonstrably pose a greater challenge for LLMs. Despite this, open-weight models performed competitively to proprietary ones. Furthermore, we found that ensembling more diverse models leads to improved scores.

In more detail, ensembling a mixture of open and closed models proved to be beneficial for factoid questions. In batch 2, single proprietary models were outperformed by such ensembling mixture. This solution also represented the best approach in batch 4, surpassing individual closed models and ensembles composed solely of open-weight or closed models. In batch 3, multiple solution types exhibited competitive performance. Table~\ref{tab:factoid} provides a summary of the results for factoid questions across all batches.

These insights are further corroborated by the findings from list questions, presented in Table~\ref{tab:list}. Ensembling solely open-weight models or a mixture of both model types consistently yielded the best approaches in batches 2, 3, and 4. For batch 1, the results between both model types were notably similar. These observations strongly suggest that ensembles of open-weight models can address more challenging tasks at a comparable, or even superior, level to closed models.

\subsection{Ideal Answers}
For the summary questions, we directly generated each summary using a chosen LLM and the prompt detailed in Table~\ref{tab:prompts}, without employing ensembling techniques. For systems involving multiple LLMs, we generated candidate summaries from all participating models for each question. The best summary was then selected using a cross-encoder reranking approach. This method involved calculating the similarity score between each generated summary and its corresponding question. The summary with the highest score was subsequently selected. For this purpose, we used the BiomedBERT model \cite{pubmedbert} to compute these similarity measures.

The recall scores for ROUGE metrics \cite{rouge} achieved by our method for batches 2 and 4 were comparable to those of the top-performing solutions.
However, F1 scores for ROUGE metrics in these batches were significantly lower. The results for summary questions can be seen in Table~\ref{tab:summary}.

As shown in Table~\ref{tab:summary}, Phi-4 exhibited the strongest performance among the evaluated LLMs, suggesting that open-weight models can also be competitive for summary-based questions. However, a full analysis of the responses to these questions requires manual scores that have not yet been published.

\begin{table*}
  \caption{Task 13B, Phase B, Yes/No questions, evaluated by exact answer matches}
  \label{tab:yesno}
  \begin{tabular}{|l|l|l|l|l|l|l|}
    \hline
    \textbf{Batch Nr}&\textbf{Position}&\textbf{System}&\textbf{Accuracy}&\textbf{F1 Yes}&\textbf{F1 No}&\textbf{Macro F1}\\
    \hline
    \multirow{5}{*}{Batch 1} 
    & 1 of 72  & EP-4 & 1.0 & 1.0 & 1.0 & \textbf{1.0} \\
    & 23 of 72 & EP-1         & 0.9412 & 0.9565 & 0.9091 & \textbf{0.9328} \\
    & & Best result & 1.0 & 1.0 & 1.0 & \textbf{1.0} \\ \hline
    
    \multirow{5}{*}{Batch 2} 
    & 1 of 72 & EP-4 & 1.0 & 1.0 & 1.0 & \textbf{1.0} \\
    & 22 of 72  & EP-5         & 0.9412 & 0.9524 & 0.9231 & \textbf{0.9377} \\
    & & Best result & 1.0 & 1.0 & 1.0 & \textbf{1.0} \\ \hline

    \multirow{5}{*}{Batch 3} 
    & 1 of 66 & EP-1 & 0.9545 & 0.9697 & 0.9091 & \textbf{0.9394} \\
    & 30 of 66 & EP-2         & 0.9091 & 0.9412 & 0.8000 & \textbf{0.8706} \\
    & & Best result & 0.9545 & 0.9697 & 0.9091 & \textbf{0.9394} \\ \hline

    \multirow{5}{*}{Batch 4} 
    & 1 of 79 & EP-1 & 1.0 & 1.0 & 1.0 & \textbf{1.0} \\
    & 1 of 79  & EP-3         & 1.0 & 1.0 & 1.0 & \textbf{1.0} \\
    & 3 of 79  & EP-4         & 0.9615 & 0.9730 & 0.9333 & \textbf{0.9532} \\
    & 3 of 79  & EP-5         & 0.9615 & 0.9730 & 0.9333 & \textbf{0.9532} \\
    & & Best result & 1.0 & 1.0 & 1.0 & \textbf{1.0} \\ \hline
\end{tabular}
\end{table*}

\begin{table*}
  \caption{Task 13B, Phase B, Factoid questions, evaluated by exact answer matches}
  \label{tab:factoid}
  \begin{tabular}{|l|l|l|l|l|l|}
    \hline
    \textbf{Batch Nr}&\textbf{Position}&\textbf{System}&\textbf{Strict Acc.}&\textbf{Lenient Acc.}&\textbf{MRR}\\
    \hline
    \multirow{5}{*}{Batch 1} 
    & 18 of 67  & EP-1 & 0.4231 & 0.5385 & \textbf{0.4808} \\
    & 18 of 67 & EP-2         & 0.4231 & 0.5385 & \textbf{0.4808} \\
    & 25 of 67  & EP-3         & 0.4231 & 0.5000 & \textbf{0.4615} \\
    & & Best result & 0.5385 & 0.6538 & \textbf{0.5962} \\ \hline
    \multirow{5}{*}{Batch 2} 
    & 10 of 66  & EP-1 & 0.5185 & 0.7407 & \textbf{0.6031} \\
    & 29 of 66  & EP-2         & 0.5185 & 0.5556 & \textbf{0.5370} \\
    & & Best result & 0.7037 & 0.7778 & \textbf{0.7037} \\ \hline

    \multirow{5}{*}{Batch 3} 
    & 1 of 59 & EP-3 & 0.4000 & 0.6500 & \textbf{0.5100} \\
    & 5 of 59  & EP-1         & 0.4000 & 0.5500 & \textbf{0.4625} \\
    & 7 of 59  & EP-2         & 0.3500 & 0.6000 & \textbf{0.4542} \\
    & 14 of 59  & EP-4         & 0.4000 & 0.4500 & \textbf{0.4250} \\
    & & Best result & 0.4500 & 0.6500 & \textbf{0.5100} \\ \hline

    \multirow{5}{*}{Batch 4} 
    & 7 of 73  & EP-3         & 0.5455 & 0.5909 & \textbf{0.5682} \\
    & 18 of 73 & EP-4 & 0.5000 & 0.5909 & \textbf{0.5455} \\
    & 18 of 73 & EP-1         & 0.5455 & 0.5455 & \textbf{0.5455} \\
    & & Best result & 0.6364 & 0.6818 & \textbf{0.6364} \\ \hline
\end{tabular}
\end{table*}

\begin{table*}
  \caption{Task 13B, Phase B, List questions, evaluated by exact answer matches}
  \label{tab:list}
  \begin{tabular}{|l|l|l|l|l|l|}
    \hline
    \textbf{Batch Nr}&\textbf{Position}&\textbf{System}&\textbf{Precision}&\textbf{Recall}&\textbf{F-Measure}\\
    \hline
    \multirow{5}{*}{Batch 1} 
    & 24 of 66 & EP-5 & 0.5583 & 0.5019 & \textbf{0.5174} \\
    & 24 of 66 & EP-3         & 0.5583 & 0.5019 & \textbf{0.5174} \\
    & & Best result & 0.6226 & 	0.6483 & \textbf{0.5327} \\ \hline
    
    \multirow{5}{*}{Batch 2} 
    & 2 of 66 & EP-1 & 0.5685 & 0.7083 & \textbf{0.6027} \\
    & 6 of 66 & EP-2         & 0.5877 & 0.5889 & \textbf{0.5721} \\
    & 10 of 66 & EP-5         & 0.6044 & 0.5415 & \textbf{0.5538} \\
    & & Best result & 0.6360 & 0.7132 & \textbf{0.6152} \\ \hline

    \multirow{5}{*}{Batch 3} 
    & 2 of 59 & EP-1 & 0.6659 & 0.6530 & \textbf{0.6331} \\
    & 6 of 59 & EP-3         & 0.5969 & 0.6787 & \textbf{0.6148} \\
    & 10 of 59 & EP-2         & 0.6364 & 0.6075 & \textbf{0.6075} \\
    & 15 of 59 & EP-4         & 0.5956 & 0.6141 & \textbf{0.5964} \\
    & & Best result & 0.6659 & 0.6787 & \textbf{0.6337} \\ \hline

    \multirow{5}{*}{Batch 4} 
    & 12 of 73  & EP-2 & 0.5709 & 0.6268 & \textbf{0.5896} \\
    & 13 of 73 & EP-4         & 0.5650 & 0.6284 & \textbf{0.5859} \\
    & 17 of 73 & EP-3         & 0.5355 & 0.6589 & \textbf{0.5768} \\
    & & Best result & 0.7491 & 0.6791 & \textbf{0.6492} \\ \hline
\end{tabular}
\end{table*}

\begin{table*}
  \caption{Task 13B, Phase B, Summary questions requiring free-form ideal answers}
  \label{tab:summary}
  \begin{tabular}{|l|l|l|l|l|l|l|}
    \hline
    \textbf{Batch Nr}&\textbf{System}&\textbf{R-2 (Rec)}&\textbf{R-2 (F1)}&\textbf{R-SU4 (Rec)}&\textbf{R-SU4 (F1)}\\
    \hline
    \multirow{5}{*}{Batch 1} 
    & EP-1 & 0.3594 & 0.2021 & 0.3585 & 0.1905 \\
    & EP-4 & 0.3488 & 0.2021 & 0.3409 & 0.2309 \\
    & Best result & 0.4726 & 0.4122 & 0.4490 & 0.4008 \\ \hline

    \multirow{5}{*}{Batch 2} 
    & EP-4 & 0.4213 & 0.2494 & 0.4165 & 0.2326 \\
    & EP-3 & 0.3816 & 0.2990 & 0.3613 & 0.2725 \\
    & Best result & 0.4838 & 0.4417 & 0.4652 & 0.4287 \\ \hline

    \multirow{5}{*}{Batch 3} 
    & EP-4 & 0.3550 & 0.2077 & 0.3545 & 0.1946 \\
    & EP-3 & 0.3340 & 0.2308 & 0.3339 & 0.2175 \\
    & Best result & 0.4309 & 0.3520 & 0.4357 & 0.3439 \\ \hline

    \multirow{5}{*}{Batch 4} 
    & EP-4 & 0.2947 & 0.2151 & 0.2973 & 0.2074 \\
    & EP-3 & 0.2931 & 0.2158 & 0.3030 & 0.2092 \\
    & Best result & 0.4139 & 0.3604 & 0.3963 & 0.3515 \\ \hline
\end{tabular}
\end{table*}

\section{Conclusion}
Our primary goal was to evaluate the competitive performance of open-weight LLMs against state-of-the-art proprietary LLMs for biomedical question answering. To this end, we rigorously tested numerous configurations, including both smaller open-weight models and various closed models. Our results consistently demonstrated the competitiveness of ensembles of open models for BioASQ 13B Phase B questions.

For yes/no questions, this thesis is supported by results from batches 2 and 3 (Table~\ref{tab:yesno}), where open-weight models outperformed closed-weight models. This trend also holds for list-based questions, with batches 2 through 4 demonstrating strong performance by open models on both factoid and list-type questions (Tables~\ref{tab:factoid} and \ref{tab:list}, respectively). For summary questions, the open-weight model Phi-4 exhibited promising performance in terms of ROUGE metrics in Batches 1 through 3, as detailed in Table~\ref{tab:summary}.

This conclusion holds significant implications. The ability to use open-weight models negates the need for proprietary solutions in every application. This is particularly relevant for applications involving highly restricted data that require on-premise deployment, a common scenario with medical data. In such contexts, smaller self-deployable models offer a compelling and practical alternative.

\begin{acknowledgments}
  This research was co-funded by the European Union – European Regional Development Fund (Programme: European Funds for a Modern Economy 2021-2027, grant no. FENG.01.01-IP.02-4479/23).
\end{acknowledgments}

\section*{Declaration on Generative AI}
  During the preparation of this work, the author(s) used Gemini in order to: Grammar and spelling check, Paraphrase and reword. After using this tool/service, the author(s) reviewed and edited the content as needed and take(s) full responsibility for the publication’s content. 

\bibliography{ep_bioasq_final}

\end{document}